\documentclass[letterpaper, 10 pt, conference]{ieeeconf}  % Comment this line out if you need a4paper
\IEEEoverridecommandlockouts                              % This command is only needed if 
\usepackage[noadjust]{cite}
\usepackage{threeparttable}
\usepackage{url}
\usepackage{xcolor}
\usepackage{kotex}
\usepackage{graphicx}
\usepackage{amsmath,amssymb,amsfonts}

\usepackage{siunitx}
\usepackage{float}
\usepackage{dsfont}
\usepackage{multicol}
\usepackage{bm}
\usepackage{booktabs}
\usepackage{tabularx}
\let\labelindent\relax
\usepackage{enumitem}
\usepackage{diagbox}
\usepackage{multirow}

\usepackage{cuted}
\usepackage{capt-of}
\usepackage[caption=false, font=footnotesize]{subfig}
\usepackage{algorithm}
\usepackage{algpseudocode}
\usepackage{textcomp}
\usepackage{array}
\usepackage{tikz}
\usetikzlibrary{patterns}
\usepackage{circledsteps}
\usepackage{tcolorbox}
\usepackage{colortbl}
\usepackage{xspace}
\tcbuselibrary{breakable, skins}
\usepackage{changepage} 
\usepackage{makecell}
\newcommand{\best}[1]{\cellcolor{gray!20}\textbf{#1}}
\newcommand{\second}[1]{\cellcolor{gray!10}{#1}}

\title{\LARGE \bf
VERIA: Verification-Centric Multimodal Instance Augmentation \\for Long-Tailed 3D Object Detection
}

\author{
Jumin Lee$^{1\ast}$\,\,\,\,\,
Siyeong Lee$^2$\,\,\,\,\,
Namil Kim$^2$\,\,\,\,\,
Sung-Eui Yoon$^1$
\\\\
$^1$KAIST\,\,\,\,\,\,\,\,\,\,\,\,\,\,\,\,\,\,$^2$Naver Labs
\thanks{$\ast$ Work done during an internship at Naver Labs.}
}

\begin{document}

\maketitle
\thispagestyle{empty}
\pagestyle{empty}
\begin{abstract}
Long-tail distributions in driving datasets pose a fundamental challenge for 3D perception, as rare classes exhibit substantial intra-class diversity yet available samples cover this variation space only sparsely.
Existing instance augmentation methods based on copy-paste or asset libraries improve rare-class exposure but are often limited in fine-grained diversity and scene-context placement.
We propose VERIA, an image-first multimodal augmentation framework that synthesizes synchronized RGB--LiDAR instances using off-the-shelf foundation models and curates them with sequential semantic and geometric verification.
This verification-centric design tends to select instances that better match real LiDAR statistics while spanning a wider range of intra-class variation.
Stage-wise yield decomposition provides a log-based diagnostic of pipeline reliability.
On nuScenes and Lyft, VERIA improves rare-class 3D object detection in both LiDAR-only and multimodal settings.
Our code is available at \url{https://sgvr.kaist.ac.kr/VERIA/}.
\end{abstract}
\section{Introduction}
\begin{figure}[t]
\begin{center}
\vspace{2mm}
\includegraphics[trim={0cm 0cm 0cm 0cm},clip,width=\linewidth]{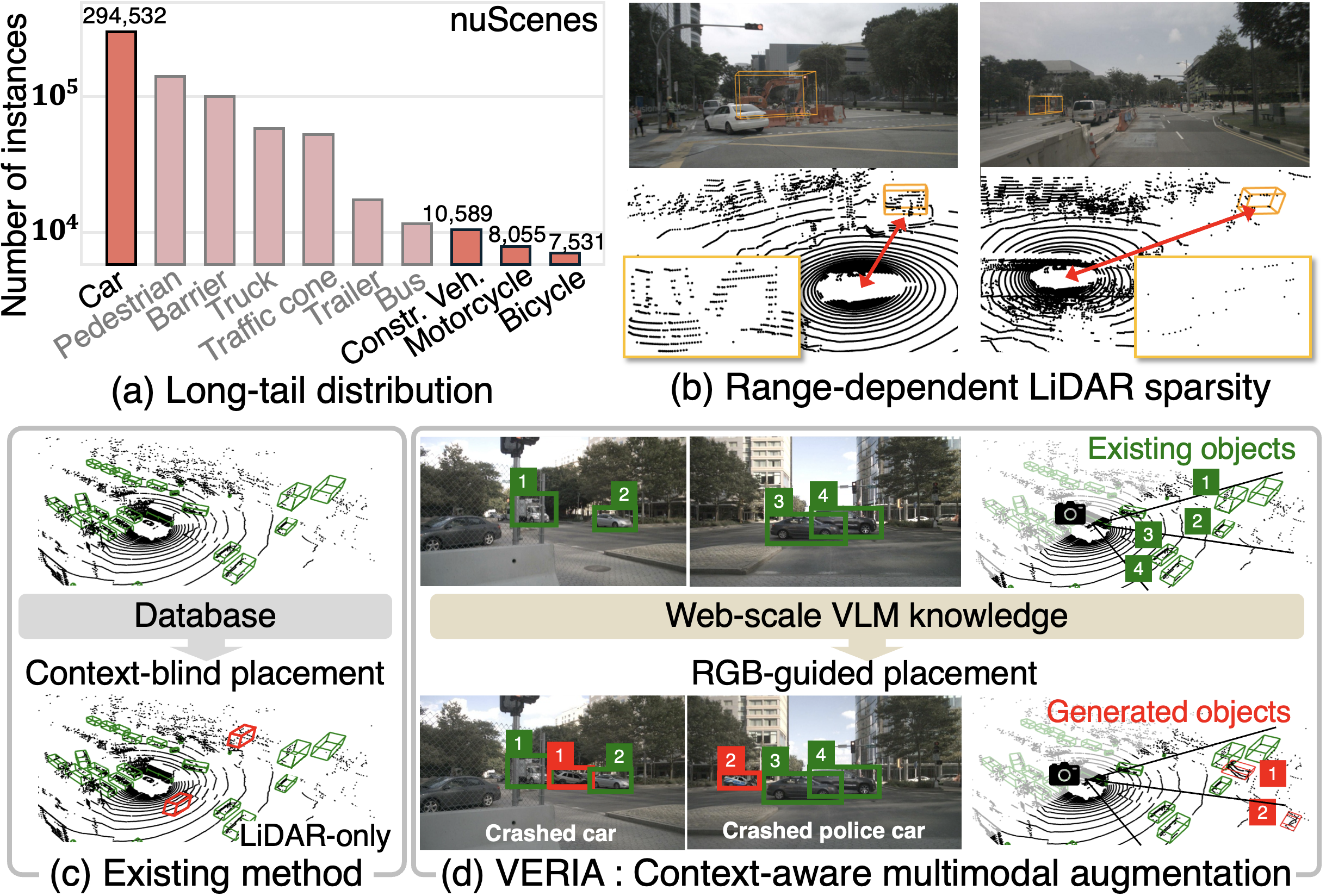}
\vspace{-7mm}
\captionsetup[subfigure]{labelformat=empty}
\caption{
\textbf{Motivation for VERIA.}
(a) Driving datasets exhibit long-tail distributions, limiting 3D perception performance.
(b) LiDAR point returns grow sparser with range, amplifying intra-class geometric variation.
(c) Existing methods operate in the LiDAR domain and place objects without scene context, constraining diversity to curated asset libraries.
(d) VERIA synthesizes objects conditioned on RGB context using foundation models, supporting subclass-level diversity with synchronized pseudo-LiDAR.}
\label{fig:longtail}
\end{center} 
\vspace{-9mm}
\end{figure}

Reliable 3D perception is a core component of autonomous driving, yet its advancement remains constrained by the long-tail distribution inherent in real-world data.
Early benchmarks such as KITTI~\cite{geiger2012kitti} and Waymo~\cite{Sun2020Waymo} circumvent this challenge by defining only three detection classes: car, pedestrian, and cyclist.
In contrast, nuScenes~\cite{caesar2020nuscenes} and Lyft~\cite{kesten2019lyft} introduce more fine-grained taxonomies, exposing severe long-tail imbalance as shown in Fig.~\ref{fig:longtail}(a).
This imbalance is not merely a matter of sample scarcity: long-tail categories often encompass multiple subcategories with substantial diversity in appearance and physical size.
This challenge is particularly pronounced for LiDAR, where point returns grow sparser with range, amplifying intra-class geometric variation.
As illustrated in Fig.~\ref{fig:longtail}(b), the same construction vehicle instance observed at close range and at distance.
Limited coverage of the resulting variation space directly degrades perception performance~\cite{zhu2019class, yaman2023instance}.

Instance augmentation has emerged as a practical remedy that increases the frequency of rare classes, through approaches ranging from copy-paste of annotated instances~\cite{yan2018second} to asset-based synthesis~\cite{reichardt2024text3daug, DBLP:conf/nips/ChangLKK24}.
While these approaches (Fig.~\ref{fig:longtail}(c)) address class imbalance, they share two fundamental limitations.
First, intra-class diversity remains bounded by the variation space of available assets, limiting coverage of rare-category appearances.
Second, placement relies on geometric feasibility, such as whether a candidate location is free of occupancy.

We address both limitations by introducing \textbf{VERIA}, an image-first instance augmentation framework, as shown in Fig.~\ref{fig:longtail}(d).
Rather than relying on constrained asset libraries, we leverage vision-language models (VLMs)~\cite{zhu2025internvl3, DBLP:journals/corr/abs-2303-08774, wu2025qwen, liu2024deepseek, dubey2024llama} to generate diverse subclass descriptions for rare categories.
We then use these descriptions to condition diffusion-based inpainting on the surrounding RGB context, synthesizing scene-consistent objects with context-consistent occlusions and diverse appearances.
We convert the synthesized RGB objects into synchronized pseudo-LiDAR via depth estimation, producing paired multimodal samples whose sampling geometry aligns with the target LiDAR.
Existing instance augmentation methods target LiDAR-only training; in contrast, VERIA produces paired RGB--LiDAR instances, enabling use in multimodal benchmarks~\cite{sindagi2019mvx, liang2022bevfusion, wu2023virtual}.

However, generation-based augmentation introduces new failure modes: synthesized objects can exhibit semantic mismatches with the intended category, and depth estimation errors can introduce geometric distortions in the reconstructed point cloud.
Without explicit filtering, such failures can enter the training set and affect downstream performance.
VERIA addresses this by treating each generated RGB--LiDAR pair as a candidate and retaining only those that pass sequential semantic and geometric verification.
This design helps mitigate error propagation while retaining a diverse set of verified candidates.
Stage-wise yield decomposition further serves as a general log-based diagnostic for multi-stage generative augmentation pipelines, quantifying where candidates are rejected across stages.

Our contributions are as follows:
\begin{itemize}
    \item We introduce VERIA, an image-first multimodal instance augmentation framework that produces context-aware RGB--LiDAR pairs for rare categories.
    %, extending instance augmentation to multimodal settings.
    \item We propose a verification-centric design with sequential semantic and geometric verification, complemented by stage-wise yield decomposition as a diagnostic of pipeline reliability.
    \item We validate VERIA on nuScenes and Lyft, demonstrating consistent rare-class 3D object detection improvements across both LiDAR-only and multimodal settings.
\end{itemize}
\vspace{-1mm}
\section{Related Work}
\subsection{Instance Augmentation for Long-Tail Problem}
Long-tail distributions remain a persistent challenge in 3D perception, and training-stage corrections such as class-balanced sampling~\cite{liu2025extremely, zhu2019class} and loss reweighting~\cite{yaman2023instance, gupta2019lvis} have been widely studied as practical remedies.
Complementary to these approaches, instance augmentation increases rare-class exposure by inserting additional instances into training scenes.
GT-Aug~\cite{yan2018second} reuses annotated objects from an instance database; however, the achievable diversity is inherently limited by database coverage.
To expand beyond fixed instance databases, PGT-Aug~\cite{DBLP:conf/nips/ChangLKK24} reconstructs 3D assets from pre-collected multi-view observations, while Text3DAug~\cite{reichardt2024text3daug} generates LiDAR assets from text prompts (e.g., a red sportscar) via text-to-mesh synthesis.

Despite these advances, diversity can remain limited by the representation and coverage of available assets, whether from pre-collected reconstructions or text-conditioned generation that does not explicitly model fine-grained subclasses.
Furthermore, placement strategies often rely on geometric constraints such as free-space or road segmentation maps, without incorporating surrounding RGB context.
Motivated by these gaps, we condition object synthesis on RGB context and leverage VLMs for subclass-level descriptions to broaden intra-class variation with context-aware placement.

\subsection{Foundation Model-Based Generative Augmentation}
Generation-based pipelines have been shown to enrich training distributions and improve downstream performance across numerous 2D perception tasks~\cite{islam2024diffusemix, kupyn2024dataset, wu2023datasetdm, yurt2025ltda, tong20243d, kimsample, hammoud2024synthclip}, suggesting foundation model-based augmentation as a promising direction in the 2D domain.
Its extension to multimodal 3D perception, however, remains less explored.
Early work addresses multi-modality augmentation through copy-paste strategies that maintain RGB-LiDAR consistency~\cite{zhang2020exploring}; however, these methods are constrained by limited instance diversity and context-blind placement that produces visually unnatural RGB scenes.
Recent work has attempted joint generation of RGB and LiDAR through inpainting conditioned on a reference RGB image~\cite{buburuzan2025mobi}; however, it requires dataset-specific training and a reference image of the target object at inference, and its effectiveness in improving 3D perception performance remains unclear.

We leverage off-the-shelf foundation models to generate synchronized RGB--LiDAR instances without additional training, and observe a improvement on rare classes across nuScenes and Lyft datasets in both LiDAR-only and multimodal settings.

\subsection{Vision-Language Model-Based Verification}
Recent advances in vision-language models~\cite{DBLP:journals/corr/abs-2303-08774, zhu2025internvl3, dubey2024llama, liu2024deepseek, wu2025qwen} have enabled verifier-based quality control for generative augmentation without additional human annotation.
Their ability to assess category consistency and visual plausibility has made them increasingly effective as annotation-free filters for retaining candidates that meet predefined criteria~\cite{lee2024prometheus, wang2025mllm, lin2025self, zhang2025vl}.
In multimodal 3D augmentation, where RGB synthesis and LiDAR reconstruction each introduce their own failure modes, annotation-free filtering across both stages becomes particularly relevant.

Building on this line of work, VERIA combines VLM-based semantic verification with geometric verification to filter errors introduced by both stages, while curating synchronized RGB--LiDAR instances.
In our evaluation, this verification-centric design is associated with lower FID and higher recall, suggesting closer resemblance to real instances and broader support over the real instance distribution.
\section{Method}\label{sec:method}
\begin{figure*}[t]
\centering
\includegraphics[trim={0cm 10.5cm 0cm 0cm},clip,width=0.9\linewidth]{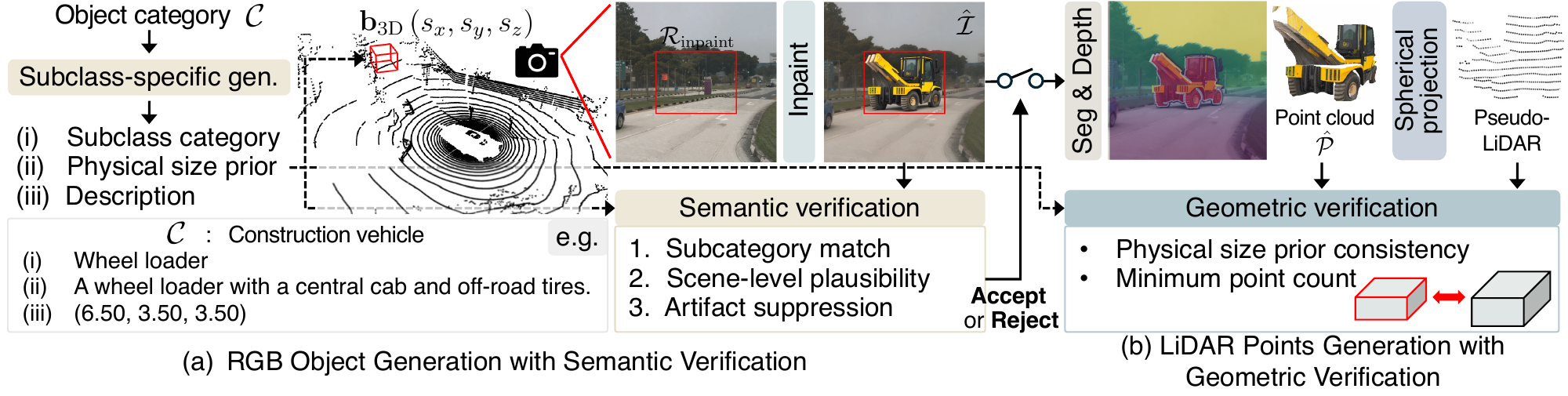}
\vspace{-4mm}
\caption{\textbf{Overview of VERIA.}
(a) Given a target category $\mathcal{C}$, a VLM generates a subclass-level description $\mathcal{T}_c$ and physical size priors; a 3D bounding box is sampled and projected to define the inpainting region for RGB-context-conditioned synthesis.
Semantic verification retains candidates that pass category correctness, scene-level plausibility, and artifact severity checks.
(b) Verified RGB instances are converted to synchronized pseudo-LiDAR via segmentation, depth estimation, and spherical projection.
Geometric verification further filters implausible reconstructions, yielding verified RGB--LiDAR pairs for downstream training.}
\label{fig:method}
\vspace{-7mm}
\end{figure*}

Our goal is to build a generative instance augmentation pipeline for long-tailed 3D object detection that increases the exposure and intra-class variation of rare categories.
The pipeline aims to maintain scene consistency in RGB and geometric plausibility in LiDAR through synchronized RGB--LiDAR instance generation and verification.
Fig.~\ref{fig:method} summarizes the overall pipeline.

\vspace{-0.5mm}
\subsection{RGB Object Generation with Semantic Verification}\label{sec:rgb_generation}
The first stage synthesizes rare-class objects in the image domain to expand intra-class diversity while preserving scene context. 
We synthesize objects via diffusion-based inpainting conditioned on the surrounding scene and VLM-derived subclass descriptions, then apply semantic verification to filter failed generations.
\begin{tcolorbox}[
  colback=gray!1,
  colframe=gray!10,
  title=Subclass Specification Prompt,
  coltitle=black,
  fonttitle=\footnotesize\bfseries,
  boxsep=1pt,
  left=1pt, right=1pt, top=1pt, bottom=1pt,
  before skip=4pt, after skip=4pt,
  %before upper=\setlength{\parskip}{0pt}\setlength{\parindent}{0pt},
  before upper={\setlength{\parskip}{0pt}\setlength{\parindent}{0pt}\spaceskip=0.5em plus 0.3em minus 0.2em},
  fontupper=\footnotesize\ttfamily,
]
Provide one subclass of \{TARGET\_LABEL\}. Include a brief visual description covering shape and notable features, along with typical physical dimensions in meters (length, width, height) as a realistic range. For dimensions, reference a concrete real-world product model and report its official specifications. If the target label is 'bicycle' or 'motorcycle', provide a description either without a rider reporting only the vehicle dimensions, or with a seated rider reporting the bounding box dimensions enclosing both the vehicle and person.
\end{tcolorbox}

\vspace{1mm}
\noindent\textbf{VLM-guided object generation in image space.}
Diffusion models and VLMs trained on web-scale data encode broad knowledge of diverse object categories.
For instance, construction vehicles encompass bulldozers, excavators, and cranes, each with distinct geometry and appearance.
We use this knowledge to generate subclass-conditioned samples that broaden the covered variation space of rare categories.

For each target category $c$, we query a VLM using the prompt above to obtain a subclass description $\mathcal{T}_c$ and physical size priors; the description serves as the inpainting condition and as the reference for semantic verification, while the size priors are used in geometric verification.

We sample candidate 3D bounding box $\mathbf{b}_{3D}=[c_x,c_y,c_z,s_x,s_y,s_z,\theta]$ where pose $(c_x,c_y,c_z,\theta)$ is drawn uniformly within the LiDAR detection range and size $(s_x,s_y,s_z)$ follows priors from $\mathcal{T}_c$.
Each box is projected onto the image plane using camera intrinsics $\mathbf{K}$ and extrinsics $[\mathbf{R}\mid\mathbf{t}]$ to form a 2D mask $\mathcal{M}$.
We extract an image patch $\mathcal{I}_{\text{patch}}$ around the inpainting region $\mathcal{R}_{\text{inpaint}}$, 
and apply an inpainting $f_{\text{in}}$~\cite{zhuang2024task} conditioned on $\mathcal{T}_c$ and $\mathcal{M}$:
\begin{equation}
    \hat{\mathcal{I}} = f_{\text{in}}\!\left(\mathcal{I}_{\text{patch}}, \mathcal{T}_c, \mathcal{M}\right).
\end{equation}

By leveraging surrounding scene cues, inpainting can produce scene-dependent appearance and orientation diversity.

\vspace{1mm}
\noindent\textbf{Semantic verification.}
We treat each synthesized image $\hat{\mathcal{I}}$ as a candidate and apply a VLM to assess sub-category correctness (Q1), scene-level plausibility in scale and placement (Q2), and local artifact severity (Q3).
To evaluate both global context and local quality, we provide the full synthesized scene with the object region marked, alongside an enlarged crop of the marked region.
The verifier additionally produces a diagnostic comment, providing interpretable feedback on each decision.

%\vspace{1mm}
\begin{tcolorbox}[
  colback=gray!1,
  colframe=gray!10,
  title=Semantic Verification Prompt,
  coltitle=black,
  fonttitle=\footnotesize\bfseries,
  boxsep=1pt,
  left=1pt, right=1pt, top=1pt, bottom=1pt,
  before skip=4pt, after skip=4pt,
  %before upper=\setlength{\parskip}{0pt}\setlength{\parindent}{0pt},
  before upper={\setlength{\parskip}{0pt}\setlength{\parindent}{0pt}\spaceskip=0.5em plus 0.3em minus 0.2em},
  fontupper=\footnotesize\ttfamily
]
You are given two images: a full driving scene with a red bounding box indicating a synthesized object region, and a cropped close-up of the boxed region. 

Q1) Does the object match the intended subclass category? (Yes/No)

Q2) Are the object's scale, placement, and orientation plausible given the surrounding scene context? (Yes/No)

Q3) How severe are visible artifacts in the object region? (none/minor/medium/severe)

Q4) Provide a brief diagnostic comment explaining your assessment.
\end{tcolorbox}
%\vspace{-1mm}
Rather than posing all criteria in a single query, which has been associated with increased hallucination rates in VLMs~\cite{li2023evaluating}, we adopt a sequential questioning strategy using the prompt above: each criterion is posed individually while accumulating prior responses in the conversation history, allowing the model to reason progressively with context from earlier assessments.
A sample passes only if Q1 and Q2 are answered Yes and Q3 is none.
Notably, Q2 evaluates object scale and placement against the full scene context, 
%providing an image-space consistency check on the VLM-derived size priors before they are applied in 3D reconstruction.
providing an image-space sanity check on scale and placement before applying size priors in 3D reconstruction.

\vspace{-0.5mm}
\subsection{LiDAR Points Generation with Geometric Verification}\label{sec:lidar_generation}
The second stage reconstructs paired pseudo-LiDAR from verified RGB objects and applies geometric verification to ensure the resulting observations are physically plausible and consistent with the synthesized scene.

\vspace{1mm}
\noindent\textbf{LiDAR points generation from monocular depth.}
We extract the generated object region $\mathcal{R}_{\text{obj}}$ using a segmentation model and predict a depth map $\hat{D}(u,v)$; the depth map is then backprojected with camera intrinsics $\mathbf{K}$ to obtain an object point cloud.
Depth discontinuities at object contours introduce mixed foreground--background pixels that manifest as boundary-induced outliers and background leakage; we suppress these by forming a narrow contour band around $\mathcal{R}_{\text{obj}}$ and removing points following MoGe2~\cite{wang2024moge}.

To resolve absolute scale ambiguity inherent in monocular depth, we normalize the reconstruction using VLM-derived size priors: letting $\hat{s}_z$ denote the vertical extent of the reconstructed object and $s_z$ the target height from $\mathcal{T}_c$, we scale the point cloud as:
\begin{equation}
\hat{\mathcal{P}}=\left\{\frac{s_z}{\hat{s}_z}\cdot \hat{D}(u,v)\,\mathbf{K}^{-1}[u,v,1]^T \,\middle|\, (u,v)\in\mathcal{R}_{\text{obj}}\right\}.
\end{equation}
Vertical height provides a relatively stable metric anchor, as the 3D-to-2D projection tends to preserve consistent vertical coverage within the inpainting region.

To align real sensor sampling, we convert $\hat{\mathcal{P}}$ to spherical coordinates, discretize it at the target sensor angular resolution, and clip it to the field of view.
The resulting range image is backprojected to obtain pseudo-LiDAR $\mathcal{P}_{\text{lidar}}$ with realistic beam density.
For nuScenes, we compute a per-point intensity map from the grayscale inpainted image, modulated by surface normal and range-based attenuation following~\cite{viswanath2024reflectivity, marcus2025synth}; 
the Lyft dataset has a constant LiDAR intensity, so reflectance simulation is omitted.

\begin{figure*}[t]
\begin{center}
    \includegraphics[trim={0cm 7cm 0cm 0cm},clip,width=0.93\linewidth]{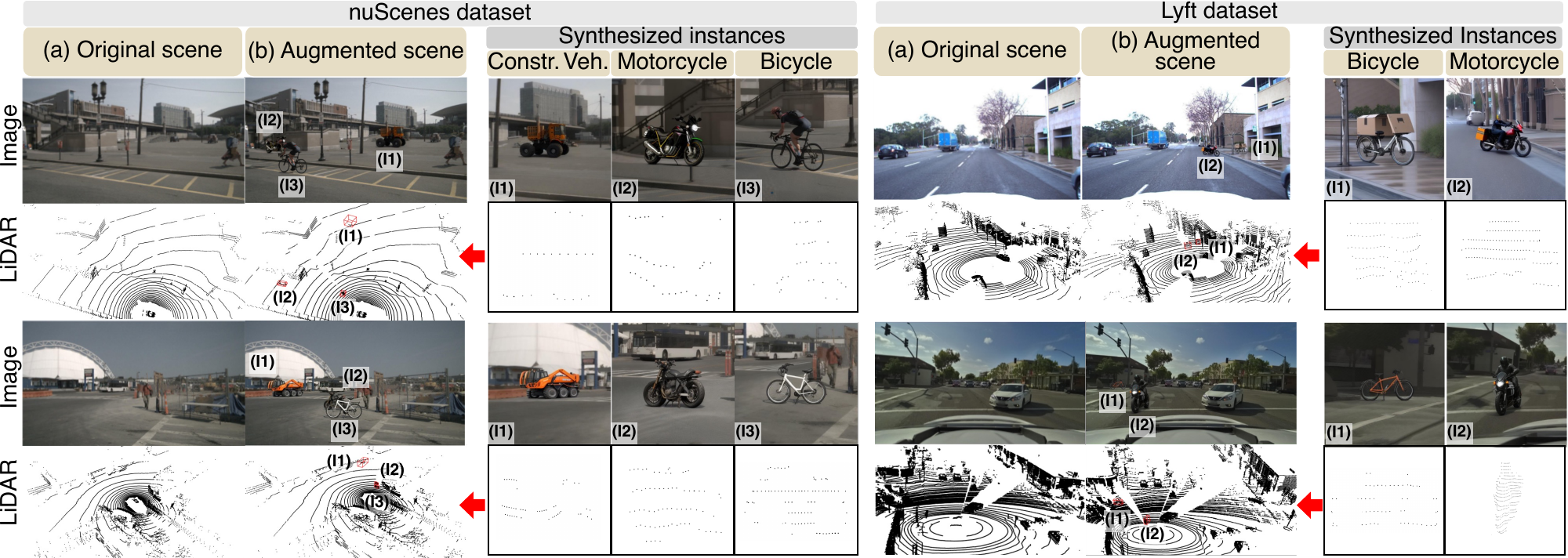}
    \vspace{-3mm}
    \caption{\textbf{Qualitative augmentations} on nuScenes and Lyft with paired RGB and LiDAR.
    For each dataset, we show an original scene (a) and the corresponding augmented scene (b), alongside the individually synthesized instances.
    On nuScenes, we augment construction vehicle (I1), motorcycle (I2), and bicycle (I3); on Lyft, we augment bicycle (I1) and motorcycle (I2).
    Each instance is composited using a collision-aware strategy; in RGB, instances are layered in depth order, while in LiDAR, occluded background points are removed from the sensor origin. 
    Red 3D bounding boxes indicate the augmented objects, which are placed at plausible locations within the scene.
    Both datasets show decreasing point density with range, consistent with real LiDAR characteristics, though Lyft's 64-beam sensor yields comparatively denser returns at distance than nuScenes' 32-beam configuration.}
    \label{fig:results}
\end{center} 
\vspace{-7mm}
\end{figure*}

\begin{figure}[t]
\begin{center}
    \includegraphics[trim={0cm 7.8cm 0.5cm 0cm},clip,width=0.85\linewidth]{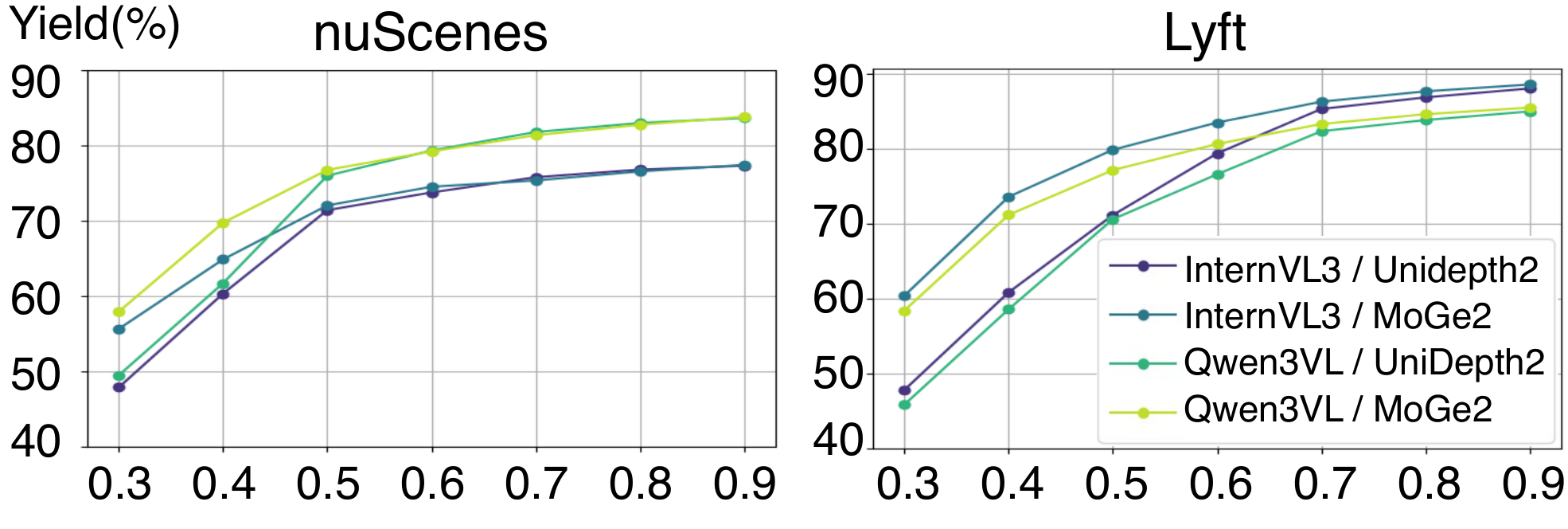}
    \vspace{-4mm}
    \caption{
    \textbf{Yield versus geometric tolerance $\lambda$} on nuScenes and Lyft. Yield increases monotonically with $\lambda$ as relaxing the consistency check admits a larger fraction of candidates.}
    \label{fig:lambda}
    \end{center} 
\vspace{-9mm}
\end{figure}

\noindent\textbf{Geometric verification.}
Even after edge-aware filtering, residual depth discontinuities can cause object points to streak behind the true surface, expanding the reconstructed bounding box along the depth axis.
We mitigate this failure mode by verifying consistency between the reconstructed geometry and the VLM-derived physical references: we fit an oriented 3D bounding box to $\hat{\mathcal{P}}$ and accept the sample only if it contains at least $p_n$ points and $(1-\lambda)s_i \le \hat{s}_i \le (1+\lambda)s_i$ for $i\in\{x,y,z\}$, where $(s_x,s_y,s_z)$ are the priors from $\mathcal{T}_c$ and $\lambda$ defines the acceptable tolerance; we set $p_n=5$ to match the OpenPCDet~\cite{openpcdet2020} database filtering criterion.

\vspace{0.5mm}
\noindent\textbf{Instance Composition.}
For each target category $\mathcal{C}$, we sample multiple candidate 3D bounding boxes with varying poses $(c_x, c_y, c_z, \theta)$ and synthesize each independently via inpainting, building a diverse instance database for each scene.
To construct training scenes with multiple synthesized instances, we employ a collision-aware composition strategy: given a base scene and this generated database, we iteratively select candidates whose 3D bounding boxes do not overlap with previously inserted objects, as illustrated in Fig.~\ref{fig:results}.
Each selected instance is composited into the segmented RGB image in depth order; occluded background LiDAR points are removed from the sensor origin, maintaining physically consistent occlusion across modalities.
For each pseudo-LiDAR $\mathcal{P}_{\text{lidar}}$, we recover a 3D bounding box $\hat{\mathbf{b}}_{\text{3D}}$ via eigen-decomposition of the point distribution in the $XY$ plane.

\begin{table}[]
\centering
\footnotesize
\setlength{\tabcolsep}{1.8pt}
\renewcommand{\arraystretch}{0.85}
\caption{\textbf{Semantic and geometric acceptance rates and final yield} over the $N$ synthesized candidates on nuScenes and Lyft, reported across four verifier--depth configurations. }
\label{tab:success_rate}
\vspace{-3mm}
\begin{tabular}{c c c c c c}
\toprule
\multirow{4}{*}{\makecell{Dataset\\(N)}} & \multirow{4}{*}{Metric}&\multicolumn{4}{c}{Method} \\
\cmidrule{3-6}
&& InternVL3 & InternVL3 & Qwen3VL & Qwen3VL \\
&& /UniDepth2 & /MoGe2 & /UniDepth2 & /MoGe2 \\
\midrule

% ==================== nuScenes ====================
\multirow{3}{*}{{\makecell{nuScenes\\(550,098)}}}
& $\mathbb{P}(S_{\text{sem}})$   & 81.29 & 81.29 & 91.71 & 91.71 \\
& $\mathbb{P}(S_{\text{geo}})$ &  81.73 & 83.60 & 81.73 & 83.60 \\
& $\mathbb{P}(S_{\text{sem}} \cap S_{\text{geo}})$     & 71.42 & 72.05 & 75.99 & 76.76 \\
\midrule
% ==================== Lyft ====================
\multirow{3}{*}{{\makecell{Lyft\\(209,270)}}}
& $\mathbb{P}(S_{\text{sem}})$   & 89.33 & 89.33 & 86.24 & 86.24 \\
& $\mathbb{P}(S_{\text{geo}})$  & 79.52 & 89.27 & 79.52 & 89.27 \\
& $\mathbb{P}(S_{\text{sem}} \cap S_{\text{geo}})$      & 71.17 & 79.87 & 70.61 & 77.16 \\
\bottomrule
\end{tabular}
\vspace{-7mm}
\end{table}

\begin{table*}[t]
\centering
\footnotesize
\setlength{\tabcolsep}{1.8pt}
\renewcommand{\arraystretch}{0.85}
\caption{\textbf{3D object detection on nuScenes.} Overall mAP and per-class AP, with augmented categories highlighted in bold. VERIA is comparable to LiDAR-only augmentation baselines and shows larger gains in the multimodal setting.}
\vspace{-3mm}
\begin{tabular}{clccccccccccc}
\toprule
\multirow{2.5}{*}{Modality} & \multirow{2.5}{*}{Method} &
\multicolumn{10}{c}{Per-class AP (\%)} & \multirow{2.5}{*}{mAP (\%)} \\
\cmidrule(lr){3-12}
& & Car & Pedestrian & Barrier & Truck & Traffic cone & Trailer & Bus & \textbf{Constr. Veh.} & \textbf{Motorcycle} & \textbf{Bicycle} & \\
\midrule
\multirow{8}{*}{\makecell{LiDAR-\\only}}
& CenterPoint &&&&&&&&&&& \\
& + GT-Aug                 & 84.86 & 85.50 & 67.40 & 58.33 & 70.77 & 39.65 & 70.59 & 21.93 & 68.67 & 57.10 & 62.48 \\
& + PGT-Aug                & 85.16 & 85.49 & 68.43 & 57.59 & 70.73 & 39.14 & 70.25 & 24.23 & 68.70 & \best{59.94} & 62.97 \\
& + Text3DAug              & 85.02 & 85.02 & 68.05 & 57.77 & 71.17 & 40.68 & 71.91 & 23.39 & 69.05 & 58.79 & 63.09 \\
& + VERIA (InternVL3/MoGe2)  & 84.68 & 85.46 & 68.44 & 59.03 & 69.99 & 40.52 & 70.23 & \second{24.33} & \best{69.71} & 58.84 & 63.12 \\
& + VERIA (Qwen3VL/MoGe2)    & 84.69 & 85.31 & 67.83 & 58.21 & 70.33 & 40.95 & 69.11 & \best{24.82} & 69.34 & 58.13 & 62.87 \\
& + VERIA (InternVL3/UniDepth2) & 84.61 & 85.31 & 68.81 & 58.71 & 70.07 & 39.74 & 70.35 & 23.17 & 69.33 & \second{59.83} & 62.99 \\
& + VERIA (Qwen3VL/UniDepth2)   & 84.87 & 84.97 & 68.39 & 59.38 & 69.80 & 40.41 & 70.04 & 24.08 & \second{69.65} & \best{59.94} & \best{63.15} \\
\midrule
\multirow{5}{*}{\makecell{Multi-\\modal}}
& BEVFusion                  & 88.12 & 87.31 & 71.38 & 57.24 & 76.92 & 40.47 & 70.63 & 28.42 & 69.90 & 54.01 & 64.44 \\
& + VERIA (InternVL3/MoGe2)     & 88.19 & 86.86 & 71.75 & 59.81 & 78.12 & 40.04 & 73.31 & 30.13 & 71.79 & 59.50 & \second{65.95} \\
& + VERIA (Qwen3VL/MoGe2)       & 88.08 & 86.72 & 72.14 & 59.69 & 78.19 & 40.50 & 72.51 & \best{31.29} & \second{72.31} & \second{60.29} & \best{66.17} \\
& + VERIA (InternVL3/UniDepth2) & 88.21 & 86.79 & 72.80 & 58.40 & 77.64 & 39.22 & 72.16 & \second{30.40} & \best{73.16} & \best{60.33} & 65.91 \\
& + VERIA (Qwen3VL/UniDepth2)   & 88.24 & 86.95 & 71.50 & 59.04 & 76.85 & 40.69 & 72.19 & 30.32 & 71.93 & 58.77 & 65.65 \\
\bottomrule
\end{tabular}
\vspace{-2mm}
\label{tab:nuscene}
\end{table*}

\begin{table*}[t] 
\centering 
\footnotesize 
\setlength{\tabcolsep}{7.0pt} 
\renewcommand{\arraystretch}{0.85} 
\caption{\textbf{3D object detection on Lyft.} Overall mAP and per-class AP, with augmented categories highlighted in bold. VERIA improves bicycle and motorcycle AP in both LiDAR-only and multimodal settings, consistent with the nuScenes results.}
\vspace{-3mm} 
\begin{tabular}{clcccccccc} 
\toprule 
\multirow{2.5}{*}{Modality} & \multirow{2.5}{*}{Method} & \multicolumn{7}{c}{Per-class AP (\%)} & \multirow{2.5}{*}{mAP (\%)} \\ 
\cmidrule(lr){3-9} & & Car & Other Veh. & Pedestrian & \textbf{Bicycle} & Truck & Bus & \textbf{Motorcycle} & \\ 
\midrule 
\multirow{7}{*}{\makecell{LiDAR-\\only}} & CenterPoint &&&&&&&& \\ & + GT-Aug & 36.40 & 30.60 & 5.80 & 5.10 & 18.60 & 20.70 & 4.40 & 17.37 \\ 
& + Text3DAug & 36.20 & 30.70 & 6.10 & 5.40 & 19.10 & 21.90 & 4.50 & 17.70 \\ 
& + VERIA (InternVL3/MoGe2) & 36.69 & 30.73 & 6.02 & 6.02 & 19.17 & 21.27 & \best{5.52} & 17.92 \\ 
& + VERIA (Qwen3VL/MoGe2) & 36.60 & 30.80 & 6.10 & 5.90 & 20.70 & 21.13 & 4.60 & \best{17.98} \\ 
& + VERIA (InternVL3/UniDepth2) & 36.69 & 31.12 & 6.03 & \best{6.26} & 19.47 & 20.86 & \second{5.12} & \second{17.94} \\ 
& + VERIA (Qwen3VL/UniDepth2) & 36.10 & 30.90 & 6.10 & \second{6.20} & 19.44 & 20.36 & 4.90 & 17.71 \\ 
\midrule 
\multirow{5}{*}{\makecell{Multi-\\modal}}
& BEVFusion                    & 40.51 & 30.34 & 4.98 & 4.85 & 18.81 & 17.72 & 1.60 & 16.97 \\
& + VERIA (InternVL3/MoGe2)     & 40.62 & 30.35 & 4.90 & 5.85 & 18.54 & 17.99 & \best{3.06} & \best{17.33} \\
& + VERIA (Qwen3VL/MoGe2)       & 40.39 & 30.88 & 5.01 & \second{6.17} & 18.71 & 17.66 & 1.94 & 17.25 \\
& + VERIA (InternVL3/UniDepth2) & 40.12 & 30.56 & 5.15 & 5.51 & 18.63 & 17.66 & 2.37 & 17.14 \\
& + VERIA (Qwen3VL/UniDepth2)   & 40.52 & 30.23 & 4.95 & \best{6.22} & 18.70 & 17.15 & \second{2.47} & \second{17.18} \\
\bottomrule 
\end{tabular} 
\vspace{-6mm} 
\label{tab:lyft} 
\end{table*}
\vspace{-1mm}
\subsection{Success-Rate Evaluation}\label{sec:verification_reliability}
The proposed pipeline consists of two generation stages and corresponding verification stages, each of which may affect the quality of the final augmented samples.
We therefore propose stage-wise yield decomposition as an annotation-free diagnostic that quantifies per-stage acceptance from generation logs, complementing downstream detection metrics.

Let $S_{\text{sem}}$ and $S_{\text{geo}}$ denote semantic and geometric verification pass events, and define the final yield $\mathbb{P}(S_{\text{sem}} \cap S_{\text{geo}})$.
Both rates are computed over the same set of $N$ candidates, enabling stage-wise diagnosis of yield differences across pipeline components.
Semantic acceptance reflects inpainting quality and context-level plausibility, including the image-space scale check via Q2, whereas geometric acceptance is driven by depth reconstruction quality.

We instantiate VERIA with two VLM verifiers and two depth estimators, resulting in four component combinations.
Specifically, we use InternVL3-14B~\cite{zhu2025internvl3} and Qwen3VL-32B~\cite{wu2025qwen} for semantic verification, and UniDepth2~\cite{piccinelli2024unidepth} and MoGe2~\cite{wang2024moge} for depth estimation.
We report all four combinations throughout the experiments to assess robustness to component choice.
As shown in Tab.~\ref{tab:success_rate}, $\mathbb{P}(S_{\text{sem}})$ is driven by the choice of semantic verifier, while $\mathbb{P}(S_{\text{geo}})$ is influenced by the depth estimator.
Accordingly, the final yield $\mathbb{P}(S_{\text{sem}} \cap S_{\text{geo}})$ varies across all four configurations.

The geometric tolerance $\lambda$ controls the strictness of the size-consistency check, with larger values admitting a broader range of reconstructed sizes, as shown in Fig.~\ref{fig:lambda}.
Since the inpainter operates on a 2D projection without explicit 3D constraints, synthesized objects can deviate from the original box dimensions; we therefore use a permissive $\lambda{=}0.5$ to accommodate plausible variation, and examine its effect on sample quality in Fig.~\ref{fig:lambda_fid}.
\section{Experiments}
\subsection{Implementation Details}
VERIA components and computational costs are summarized in Tab.~\ref{tab:cost}.
We report results for all four combinations of two semantic verifiers (InternVL3-14B, Qwen3VL-32B) and two depth estimators (UniDepth2, MoGe2) to assess robustness to component choice.
Although Tab.~\ref{tab:cost} reports wall-clock time per image for clarity, we run the pipeline with a batch size of 16 across 8 A100 GPUs.
For semantic verification, all VLMs are run with a fixed random seed of 42 and \texttt{max\_new\_tokens} set to 512.
Responses are parsed into structured JSON and evaluated against a deterministic decision rule, ensuring reproducibility across runs.

We evaluate on 3D object detection using OpenPCDet~\cite{openpcdet2020}, following the default training settings of Text3DAug~\cite{reichardt2024text3daug} and PGT-Aug~\cite{DBLP:conf/nips/ChangLKK24}, including the augmentation disabling hook in the final epochs to allow fine-tuning on the original data distribution.
On nuScenes~\cite{caesar2020nuscenes}, we train CenterPoint~\cite{yin2021center} for 20 epochs and BEVFusion~\cite{liang2022bevfusion} for 6 epochs, augmenting \textit{construction vehicle}, \textit{motorcycle}, and \textit{bicycle} with up to seven, five, and five instances per scene, respectively, following GT-Aug~\cite{yan2018second}.
On Lyft~\cite{kesten2019lyft}, we train CenterPoint for 30 epochs and BEVFusion for 12 epochs, augmenting \textit{motorcycle} and \textit{bicycle}, which exhibit the lowest detection performance among less frequent classes, with up to six instances per scene; for BEVFusion, we follow the nuScenes image processing configuration as no official configuration is provided in OpenPCDet. 
As PGT-Aug is not publicly available, Lyft comparisons are limited to GT-Aug and Text3DAug.
Following PGT-Aug, emergency vehicle and animal are excluded, as the validation split contains only one unique emergency vehicle and two unique animal instances spanning near-duplicate frames, rendering per-class AP an unreliable evaluation signal.
\begin{table}[]
\centering
\footnotesize
\vspace{2mm}
\caption{\textbf{Model and inference time} (s) measured on an A100 GPU.}
\label{tab:cost}
\vspace{-3mm}
\setlength{\tabcolsep}{1.8pt}
\renewcommand{\arraystretch}{0.9}
\begin{tabular}{cccc}
\toprule
Module & {Model} & {Time(s)} \\
\midrule
Inpainter & PowerPaint~\cite{zhuang2024task}  & 1.08 \\
VLM & InternVL3-14B~\cite{zhu2025internvl3} / Qwen3VL-32B~\cite{wu2025qwen}    &  2.18 / 2.36 \\
Segmentation & SAM2~\cite{ravi2024sam}   & 0.14 \\
Depth estimator& MoGe2~\cite{wang2024moge} / UniDepth2~\cite{piccinelli2024unidepth} & 0.39 / 0.37\\
\bottomrule
\end{tabular}
\vspace{-6mm}
\end{table}

\begin{figure*}[t]
\centering
    \centering
    \includegraphics[trim={0cm 14.5cm 0cm 0cm},clip,width=0.97\linewidth]{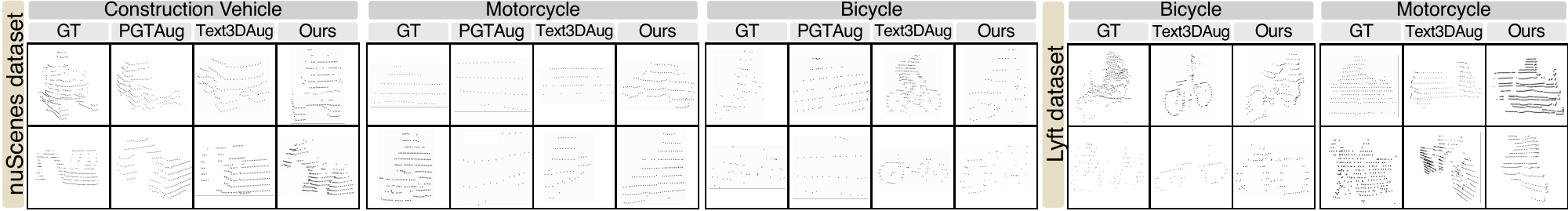}
    \vspace{-3mm}
    \captionof{figure}{\textbf{Qualitative pseudo-LiDAR comparison.} VERIA against PGT-Aug and Text3DAug using MoGe2 (first row) and UniDepth2 (second row). 
    All visualized instances contain at least 64 points. 
    Despite relying on depth-based reconstruction, VERIA produces beam-pattern-consistent pseudo-LiDAR comparable to mesh-based outputs on nuScenes (32-beam) and Lyft (64-beam).
    Mesh-based instances can appear visually cleaner than real scans, as they often lack sensor noise and irregular returns; Tab.~\ref{tab:quality} provides complementary quantitative evaluation.}
    \label{fig:quality}
    \vspace{-6mm}
\end{figure*}

\begin{table}[t]
\footnotesize
\vspace{2mm}
\setlength{\tabcolsep}{1.8pt}
\renewcommand{\arraystretch}{0.85}
\caption{\textbf{Pseudo-LiDAR instance quality on nuScenes}, measured by FID for distributional fidelity and $k$-NN precision and recall for sample-level authenticity and diversity.}
\label{tab:quality}
\vspace{-3mm}
\begin{tabular}{l l c c c c}
\toprule
Class & Method & FID $\downarrow$ & Precision $\uparrow$ & Recall $\uparrow$ & F1 $\uparrow$ \\
\midrule
\multirow{6}{*}{\makecell{Constr.\\Veh.}}
& PGT-Aug            & 2.96 & \best{0.95} & 0.81 & 0.87 \\
& Text3dAug          & 2.54 & 0.92        & 0.87 & 0.89 \\
& VERIA\,(InternVL3/UniDepth2) & 2.15 & 0.94        & \best{0.89} & \best{0.91} \\
& VERIA\,(InternVL3/MoGe2)     & 2.17 & 0.94        & 0.85 & 0.89 \\
& VERIA\,(Qwen3VL/UniDepth2)   & \best{2.12} & 0.94 & 0.87 & \second{0.90} \\
& VERIA\,(Qwen3VL/MoGe2)       & \second{2.16} & 0.94 & 0.83 & 0.88 \\
\midrule

\multirow{6}{*}{\makecell{Motor-\\cycle}}
& PGT-Aug            & 1.51 & \best{0.93} & 0.85 & 0.88 \\
& Text3dAug          & 4.72 & 0.89        & 0.87 & 0.88 \\
& VERIA\,(InternVL3/UniDepth2) & 1.28 & \second{0.92} & 0.89 & 0.90 \\
& VERIA\,(InternVL3/MoGe2)     & 1.27 & 0.90        & 0.89 & 0.90 \\
& VERIA\,(Qwen3VL/UniDepth2)   & \second{1.27} & 0.90 & 0.88 & 0.89 \\
& VERIA\,(Qwen3VL/MoGe2)       & \best{1.25} & 0.90        & 0.89 & 0.90 \\
\midrule

\multirow{6}{*}{\makecell{Bi-\\cycle}}
& PGT-Aug            & 3.32 & \best{0.92} & 0.79 & 0.85 \\
& Text3dAug          & 9.97 & 0.90        & 0.66 & 0.76 \\
& VERIA\,(InternVL3/UniDepth2) & \best{3.21} & 0.88 & \best{0.91} & 0.89 \\
& VERIA\,(InternVL3/MoGe2)     & 3.27 & 0.88        & 0.89 & 0.89 \\
& VERIA\,(Qwen3VL/UniDepth2)   & \second{3.23} & 0.89 & 0.89 & 0.89 \\
& VERIA\,(Qwen3VL/MoGe2)       & 3.24 & 0.90        & 0.89 & \best{0.90} \\
\bottomrule
\end{tabular}
\vspace{-6mm}
\end{table}

\subsection{3D Object Detection}
Tab.~\ref{tab:nuscene} and Tab.~\ref{tab:lyft} report nuScenes and Lyft results for both LiDAR-only and multimodal detectors, averaged over three runs.
On nuScenes, VERIA improves AP on the three augmented categories across the four instantiations, while maintaining comparable accuracy on frequent classes.
In the LiDAR-only setting, VERIA is comparable to LiDAR-only augmentation baselines such as PGT-Aug and Text3DAug, supporting depth-based pseudo-LiDAR reconstruction as a practical alternative to mesh-asset generation.
The gains are more pronounced in the multimodal setting, where VERIA provides synchronized RGB--LiDAR instances that fusion models can directly use; most existing instance augmentation methods target LiDAR-only training and are not directly applicable in this setting.
On Lyft, VERIA improves motorcycle and bicycle AP in both LiDAR-only and multimodal settings, suggesting similar behavior across datasets.
%The consistent improvement across component choices aligns with verification-centric selection contributing to rare-class detection gains.
Consistent gains across component choices suggest that verification-centric selection contributes to rare-class detection improvements.

\vspace{-1mm}
\subsection{Instance Quality Evaluation}
\vspace{-0.5mm}
We evaluate the quality of synthesized LiDAR instances following the PGT-Aug protocol, extracting instance embeddings using an SE(3)-Transformer~\cite{fuchs2020se} trained on nuScenes.
We report FID, $k$-NN precision and recall with $k{=}5$, and their harmonic mean (F1) per category for instances with at least 64 points.
FID measures overall distributional similarity between synthesized and real instances in the embedding space, while precision and recall are computed from nearest-neighbor relationships between individual embeddings, capturing sample-level authenticity and variation space coverage, respectively.
As shown in Tab.~\ref{tab:quality} and Fig.~\ref{fig:quality}, VERIA variants achieve lower FID than PGT-Aug and Text3DAug across all categories and model combinations.
Recall improves consistently across the four instantiations, while precision remains comparable to the baselines.
The slightly lower precision compared to PGT-Aug is consistent with increased intra-class diversity from subclass-level synthesis, which can generate instances in less frequent regions of the real embedding distribution.
As a result, VERIA achieves higher F1 than both baselines.

\vspace{-0.5mm}
\subsection{Ablations}
\vspace{-0.5mm}
\noindent\textbf{Ablation of semantic verification.}
We evaluate semantic verification outcomes using CLIP (ViT-B/32)~\cite{hessel2021clipscore}, grouping samples into Pass, Fail (Category), and Fail (Scale).
Median CLIPScore and interquartile range are reported in Tab.~\ref{tab:clip_ablation} and Tab.~\ref{tab:iqa_ablation_avg}.
CLIPScore is computed as $100\cdot\max(0,\cos)$ between image embeddings and two prompt sets: category prompts such as ``a photo of a \{class\}'' and scale prompts such as ``a street scene with a \{class\} at a realistic size and perspective.''
This serves as an annotation-free proxy for whether verifier-induced groupings correspond to meaningful differences in visual content.
Across both nuScenes and Lyft, Pass achieve the highest CLIPScores, while Fail (Category) is most penalized by category prompts and Fail (Scale) by scale prompts, suggesting that verifier-induced groupings capture meaningful differences in image-level alignment.
\begin{table}[t]
\vspace{2mm}
\centering
\footnotesize
\setlength{\tabcolsep}{2.4pt}
\renewcommand{\arraystretch}{0.9}
%\caption{{Image-level CLIPScore ablation on nuScenes and Lyft.} }
\caption{\textbf{Image-level CLIPScore} ablation by semantic verification.}%Accepted samples achieve the highest CLIPScore; Fail (Category) is lowest under category prompts, and Fail (Scale) under scale prompts.}
\vspace{-3mm}
\label{tab:clip_ablation}
\begin{tabular}{c c c cc}
\toprule
\makecell{Data} & VLM & Sem. Verif. &
\makecell{Category\\CLIP $\uparrow$ [IQR]} &
\makecell{Scale\\CLIP $\uparrow$ [IQR]} \\
\midrule

% ==================== nuScenes ====================
\multirow{6}{*}{\rotatebox[origin=c]{90}{nuScenes}}
& \multirow{3}{*}{\rotatebox[origin=c]{90}{\makecell{Intern\\VL3}}}
& PASS            & \best{26.28 [24.30, 27.71]} & \best{28.83 [27.09, 30.28]} \\
& 
& FAIL (Category) & 24.23 [22.80, 25.54]        & \second{27.86 [25.57, 29.75]} \\
&
& FAIL (Scale)    & \second{25.53 [23.02, 27.66]} & 27.04 [25.49, 28.43] \\
\cmidrule(lr){2-5}
& \multirow{3}{*}{\rotatebox[origin=c]{90}{\makecell{Qwen\\3VL}}}
& PASS            & \best{26.29 [24.25, 27.75]} & \best{28.84 [27.05, 30.30]} \\
&
& FAIL (Category) & 24.49 [22.91, 25.82]        & \second{27.18 [25.59, 28.58]} \\
&
& FAIL (Scale)    & \second{25.14 [22.63, 27.36]} & 27.10 [24.71, 29.08] \\
\midrule

% ==================== Lyft ====================
\multirow{6}{*}{\rotatebox[origin=c]{90}{Lyft}}
& \multirow{3}{*}{\rotatebox[origin=c]{90}{\makecell{Intern\\VL3}}}
& PASS            & \best{25.28 [23.26, 27.00]} & \best{29.36 [27.78, 30.82]} \\
&
& FAIL (Category) & 23.97 [22.66, 25.40]        & \second{28.44 [27.05, 29.67]} \\
&
& FAIL (Scale)    & \second{24.81 [22.72, 27.03]} & 28.32 [26.46, 30.09] \\
\cmidrule(lr){2-5}
& \multirow{3}{*}{\rotatebox[origin=c]{90}{\makecell{Qwen\\3VL}}}
& PASS            & \best{25.35 [23.30, 27.07]} & \best{29.40 [27.81, 30.85]} \\
&
& FAIL (Category) & 24.10 [22.40, 25.89] & \second{28.53 [27.14, 29.79]} \\
&
& FAIL (Scale)    & \second{24.31 [22.83, 26.16]}        & 27.92 [26.18, 29.65] \\
\bottomrule
\end{tabular}
\vspace{-3mm}
\end{table}

\begin{table}[t]
\centering
\footnotesize
\setlength{\tabcolsep}{3pt}
\renewcommand{\arraystretch}{1.0}
\caption{\textbf{NIQE\,and\,BRISQUE\,scores}\,ablation\,by\,semantic\,verification.}
%Accepted samples yield lower NIQE and BRISQUE scores than Fail (Artifact) samples.}
\vspace{-3mm}
\label{tab:iqa_ablation_avg}
\begin{tabular}{c c c cc}
\toprule
\makecell{Dataset} & VLM & Sem. Verif. &
\makecell{NIQE $\downarrow$ [IQR]} &
\makecell{BRISQUE $\downarrow$ [IQR]} \\
\midrule

% ==================== nuScenes ====================
\multirow{4.5}{*}{\rotatebox[origin=c]{90}{nuScenes}}
%& \multicolumn{2}{c}{All samples} & { [, ]} & { [, ]} \\
%\cmidrule(lr){2-5}
& \multirow{2}{*}{\rotatebox[origin=c]{90}{\makecell{Intern\\VL3}}}
& PASS            & \best{3.99 [3.63, 4.39]} & \best{28.26 [23.07, 33.79]} \\
&
& FAIL (Artifact) & {4.10 [3.72, 4.51]} & {30.00 [24.46, 35.72]} \\
\cmidrule(lr){2-5}
& \multirow{2}{*}{\rotatebox[origin=c]{90}{\makecell{Qwen\\3VL}}}
& PASS            & \best{3.99 [3.62, 4.39]} & \best{28.15 [22.95, 33.69]} \\
&
& FAIL (Artifact) & {4.16 [3.72, 4.72]} & {30.41 [25.18, 36.42]} \\
\midrule

% ==================== Lyft ====================
\multirow{4.5}{*}{\rotatebox[origin=c]{90}{Lyft}}
%& \multicolumn{2}{c}{All samples} & { [, ]} & { [, ]} \\
%\cmidrule(lr){2-5}
& \multirow{2}{*}{\rotatebox[origin=c]{90}{\makecell{Intern\\VL3}}}
& PASS            & \best{3.85 [3.40, 4.35]} & \best{25.15 [20.00, 30.73]} \\
&
& FAIL (Artifact) & {4.06 [3.53, 4.66]} & {25.76 [20.36, 31.99]} \\
\cmidrule(lr){2-5}
& \multirow{2}{*}{\rotatebox[origin=c]{90}{\makecell{Qwen\\3VL}}}
& PASS            & \best{3.85 [3.40, 4.35]} & \best{25.05 [19.91, 30.61]} \\
&
& FAIL (Artifact) & {4.21 [3.62, 4.84]} & {26.73 [21.22, 33.37]} \\
\bottomrule
\end{tabular}
\vspace{-7mm}
\end{table}

\begin{figure*}[t]
\centering
\includegraphics[trim={0cm 9cm 0cm 0cm},clip,width=0.95\linewidth]{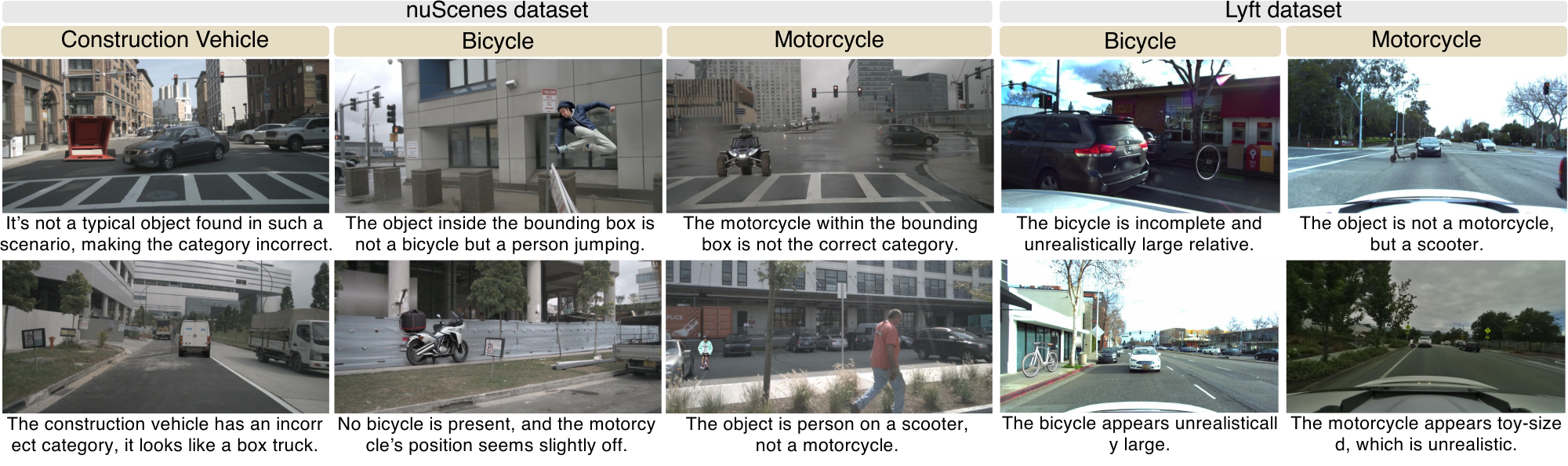}
\vspace{-4mm}
\caption{\textbf{Examples rejected by semantic verification on nuScenes and Lyft.}
Representative samples filtered by the semantic verifier across categories, with the verifier's diagnostic comment shown below each image.
Rejections include category mismatches, implausible scale or placement, and artifacts.}
\label{fig:fail}
\vspace{-2mm}
\end{figure*}
\begin{table*}[t]
    \centering
    \footnotesize
    \setlength{\tabcolsep}{2.8pt}
    \renewcommand{\arraystretch}{0.85}
    \caption{\textbf{3D object detection ablation on nuScenes.} Augmented-category AP and mAP across semantic and geometric verification configurations, showing how verification choices affect downstream performance. }%Full verification generally yields the highest scores.}
    \label{tab:ablation}
    \vspace{-3mm}
    \begin{tabular}{llcccc|cccc|cccc|cccc}
    \toprule
    &  & \multicolumn{12}{c}{Per-class AP (\%)} & \\
    \cmidrule{3-14}
    \multirow{-2.5}{*}{Modality} & \multirow{-2.5}{*}{Method} & \multicolumn{4}{c}{Constr. Veh.} & \multicolumn{4}{c}{Motorcycle} & \multicolumn{4}{c}{Bicycle} & \multicolumn{4}{c}{\multirow{-2.5}{*}{mAP (\%)}} \\
    \midrule
    & Sem. Verif. & \checkmark & --         & \checkmark & -- & \checkmark & --         & \checkmark & -- & \checkmark & --         & \checkmark & -- & \checkmark & --         & \checkmark & -- \\
    & Geo. Verif. & \checkmark & \checkmark & --         & -- & \checkmark & \checkmark & --         & -- & \checkmark & \checkmark & --         & -- & \checkmark & \checkmark & --         & -- \\
    \midrule
    \multirow{4}{*}{\makecell{LiDAR-only\\(CenterPoint)}}
    &InternVL3/MoGe2      & \best{24.33} & \second{23.14} & 23.12 & 22.79 & \best{69.71} & \second{69.35} & 67.29 & 67.29 & \best{58.84} & \second{57.65} & 57.47 & 57.41 & \best{63.12} & \second{62.96} & 62.74 & 62.43 \\
    &Qwen3VL/MoGe2        & \best{24.82} & \second{22.82} & 22.40 & 22.10 & \best{69.54} & \second{69.33} & 69.06 & 67.62 & \best{58.13} & \second{57.48} & 57.38 & 55.92 & \second{62.90} & \best{62.87} & 62.49 & 62.39 \\
    &InternVL3/UniDepth2  & \best{23.17} & \second{22.89} & 22.88 & 22.27 & \best{69.33} & \second{69.29} & 67.89 & 67.89 & \best{59.83} & \second{58.53} & 57.34 & 56.09 & \best{62.99} & \second{62.75} & 62.35 & 62.48 \\
    &Qwen3VL/UniDepth2    & \best{24.08} & \second{23.71} & 23.54 & 22.75 & \best{69.65} & \second{69.52} & 68.71 & 68.72 & \best{59.94} & \second{57.58} & 57.41 & 56.69 & \best{63.15} & {62.78} & {62.78} & 62.53 \\
    \midrule
    \multirow{4}{*}{\makecell{Multimodal\\(BEVFusion)}}
    &InternVL3/MoGe2      & \best{30.13} & \second{29.36} & 28.69 & 28.36 & \best{71.79} & {71.59} & {71.59} & 70.07 & \best{59.50} & \second{58.59} & 57.71 & 57.79 & \best{65.95} & \second{65.19} & 65.13 & 64.99 \\
    &Qwen3VL/MoGe2        & \best{31.29} & \second{29.36} & 29.03 & 28.42 & \best{72.31} & \second{72.18} & 70.49 & 70.09 & \best{60.29} & \second{58.59} & 58.47 & 57.55 & \best{65.88} & \second{65.25} & 65.23 & 64.83 \\
    &InternVL3/UniDepth2  & \best{30.40} & \second{28.04} & 28.02 & 28.02 & \best{73.16} & \second{72.16} & 71.94 & 69.25 & \best{60.33} & \second{58.97} & \second{58.97} & 58.03 & \best{65.72} & \second{65.37} & 62.38 & 65.28 \\
    &Qwen3VL/UniDepth2    & \best{30.32} & \second{28.89} & 28.60 & 28.60 & \best{71.93} & \second{71.74} & 71.72 & 70.35 & \best{58.77} & \second{58.13} & 58.10 & 57.71 & \best{65.29} & \second{65.24} & 65.22 & 64.46 \\
    \bottomrule
    \end{tabular}
    \vspace{-6mm}
\end{table*}
\begin{table}[t]
\centering
\vspace{2mm}
\footnotesize
\setlength{\tabcolsep}{2.8pt}
\renewcommand{\arraystretch}{0.9}
\caption{\textbf{FID ablation of point-cloud instances on nuScenes.} FID under semantic and geometric verification configurations.}
\label{tab:fid_ablation}
\vspace{-3mm}
\begin{tabular}{l l c c c c}
\toprule
\multirow{3}{*}{} & \multirow{3}{*}{}  & \multicolumn{4}{c}{FID $\downarrow$} \\
\cmidrule(lr){3-6}
Class & Method & Sem.\ \checkmark & Sem.\ $\times$ & Sem.\ \checkmark & Sem.\ $\times$ \\
& & Geo.\ \checkmark & Geo.\ \checkmark & Geo.\ $\times$   & Geo.\ $\times$ \\
\midrule
\multirow{4}{*}{Constr. Veh.}
& InternVL3/UniDepth2 & \best{2.15} & \second{2.16} & 2.18 & 2.22 \\
& InternVL3/MoGe2     & \best{2.17} & 2.19 & \second{2.18} & 2.32\\
& Qwen3VL/UniDepth2   & \best{2.12} & \second{2.19} & 2.21 & 2.22 \\
& Qwen3VL/MoGe2       & \best{2.16} & \second{2.16} & {2.21} & 2.31 \\
\midrule
\multirow{4}{*}{Motorcycle}
& InternVL3/UniDepth2 & \best{1.28} & \second{1.29} & 1.30 & 1.42 \\
& InternVL3/MoGe2     & \second{1.27} & \best{1.21} & 1.28 & 1.51 \\
& Qwen3VL/UniDepth2   & \best{1.27} & {1.29} & \second{1.28} & 1.40  \\
& Qwen3VL/MoGe2       & \best{1.25} & \second{1.26} & 1.27 & {1.27} \\
\midrule
\multirow{4}{*}{Bicycle}
& InternVL3/UniDepth2 & \best{3.21} & \second{3.26} & 3.35 & 3.46 \\
& InternVL3/MoGe2     & \best{3.27} & \second{3.28} & 3.35 & 3.47 \\
& Qwen3VL/UniDepth2   & \best{3.23} & \second{3.25} & 3.35 & 3.98 \\
& Qwen3VL/MoGe2       & \best{3.24} & \second{3.30} & 3.34 & 3.52 \\
\bottomrule
\end{tabular}
\vspace{-4mm}
\end{table}

\begin{figure}[t]
\begin{center}
\vspace{1mm}
\includegraphics[trim={0.0cm 10.5cm 0cm 0cm},clip,width=\linewidth]{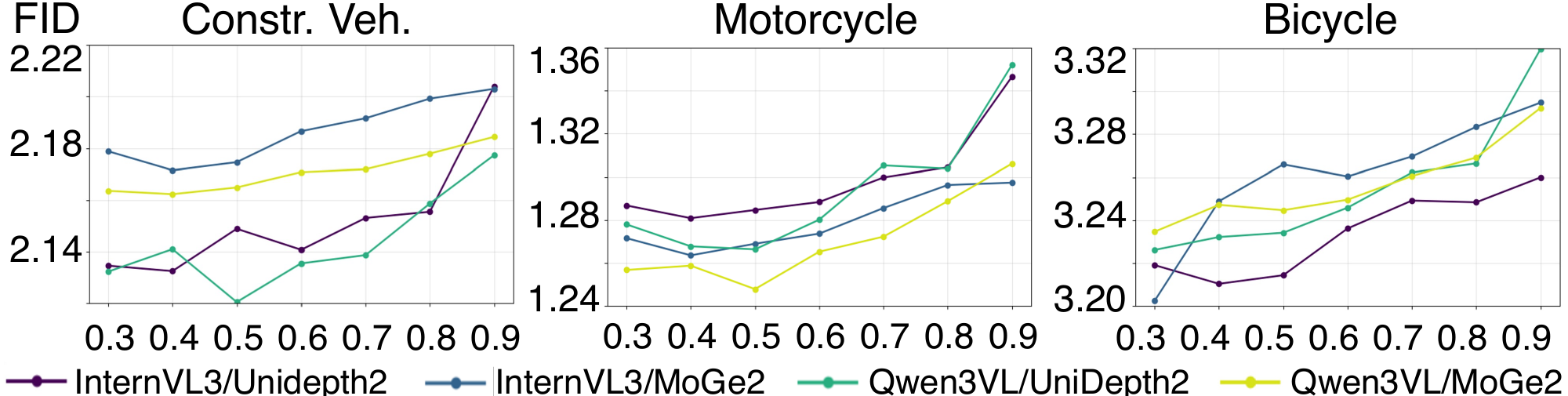}
\vspace{-7mm}
\caption{\textbf{FID vs. geometric tolerance $\lambda$ on nuScenes.} FID is lowest at moderate $\lambda$ and generally increases as $\lambda$ becomes more permissive; overly strict tolerances can also raise FID.}
\label{fig:lambda_fid}
\end{center} 
\vspace{-9mm}
\end{figure}

We further validate the artifact criterion through no-reference image quality assessment: Pass samples consistently yield lower NIQE and BRISQUE scores than Fail (Artifact) in Tab.~\ref{tab:iqa_ablation_avg}, confirming that the artifact criterion captures perceptible image degradation.
Fig.~\ref{fig:fail} illustrates representative rejected samples alongside the verifier's diagnostic comments, suggesting that the rejection criteria capture perceptually meaningful distinctions.

\vspace{1mm}
\noindent\textbf{Ablation of verification via point-cloud FID.}
Tab.~\ref{tab:fid_ablation} examines how semantic and geometric verification affect point-cloud instance FID in the SE(3)-embedding space.
FID is lowest when samples pass both stages and increases when either stage is removed.
The Sem.~$\times$ / Geo.~\checkmark setting often yields lower FID than Sem.~\checkmark / Geo.~$\times$, suggesting a more pronounced contribution from geometric verification to embedding-level similarity.

As shown in Fig.~\ref{fig:lambda_fid}, FID generally increases with $\lambda$, suggesting that a more relaxed size-consistency check admits samples with lower embedding-level fidelity; at the same time, overly strict tolerances can reduce diversity, which may also increase FID by narrowing the generated distribution.

\vspace{0.5mm}
\noindent\textbf{Ablation of verification via 3D object detection.}
To assess the impact on downstream performance, Tab.~\ref{tab:ablation} reports rare-class AP and mAP for CenterPoint and BEVFusion across verification configurations.
In the LiDAR-only setting, full verification consistently yields rare-class AP gains, while disabling both stages reduces performance below the GT-Aug baseline, 
indicating that verification helps improve augmentation reliability.
In the multimodal setting, even unverified augmentation remains above the baseline; yet enabling both stages brings additional gain, suggesting that verification improves the consistency of cross-modal supervision.

\vspace{0.5mm}
\noindent\textbf{Discussion.}
Although VLM hallucinations~\cite{li2023evaluating} and the sim-to-real gap~\cite{viswanath2024reflectivity, marcus2025synth} are inherent to generation-based augmentation, consistent gains across configurations suggest that verification-centric design helps limit their practical impact.
Prior work further reports limited downstream sensitivity to intensity simulation~\cite{DBLP:conf/nips/ChangLKK24}.
VERIA is modular in the sense that inpainting, verification, and depth estimation components can be updated independently as foundation models improve.
Our evaluation is on nuScenes and Lyft, which have limited coverage of extreme cases such as severely damaged vehicles; broader evaluation in such scenarios remains future work.
\section{Conclusion}
We presented VERIA, a multimodal augmentation pipeline for long-tail 3D object detection.
By grounding instance synthesis in the RGB domain, VERIA enables subclass-conditioned generation with context-aware placement and extends to multimodal settings via synchronized pseudo-LiDAR construction.
Sequential semantic and geometric verification helps mitigate failure modes, and stage-wise yield decomposition provides a practical diagnostic for pipeline reliability.
On nuScenes and Lyft, VERIA improves rare-class detection in both LiDAR-only and multimodal settings, alongside synthesized instances that resemble real samples while covering a broader range of intra-class variation.

{
    \small
    \bibliographystyle{IEEEtran}
    \bibliography{main}    
}

\end{document}